\documentstyle[aclap]{article}

\title{\vspace{-0.5in}Software Infrastructure for Natural Language
Processing}
\author{
\small
Hamish Cunningham \\
\small
Dept.~Computer Science \\
\small
University of Sheffield \\
\small
211 Portobello St. \\
\small
Sheffield S10 4DP \\
\small
{\tt hamish@dcs.shef.ac.uk}\And
\small
Kevin Humphreys \\
\small
Dept.~Computer Science \\
\small
University of Sheffield \\
\small
211 Portobello St. \\
\small
Sheffield S10 4DP \\
\small
{\tt kwh@dcs.shef.ac.uk}\And
\small
Robert Gaizauskas \\
\small
Dept.~Computer Science \\
\small
University of Sheffield \\
\small
211 Portobello St. \\
\small
Sheffield S10 4DP \\
\small
{\tt robertg@dcs.shef.ac.uk}\And
\small
Yorick Wilks \\
\small
Dept.~Computer Science \\
\small
University of Sheffield \\
\small
211 Portobello St. \\
\small
Sheffield S10 4DP \\
\small
{\tt yorick@dcs.shef.ac.uk}
\normalsize
}

\begin{document}
\bibliographystyle{fullname}
\maketitle
\vspace{-0.5in}

\begin{abstract}
  We classify and review current approaches to software
  infrastructure for research, development and delivery
  of NLP systems. The task is motivated by a discussion of current trends in
  the field of NLP and Language Engineering.
  We describe a system called GATE (a General Architecture for Text
  Engineering) that provides a software
  infrastructure on top of which heterogeneous NLP processing modules may be
  evaluated and refined individually, or may be combined into larger
  application systems. GATE aims to support both researchers and
  developers working on component technologies (e.g.\ parsing, tagging,
  morphological analysis) and those working on developing end-user
  applications (e.g.\ information extraction, text summarisation,
  document generation, machine translation, and second language
  learning). GATE promotes reuse of component
  technology, permits specialisation and collaboration in large-scale
  projects, and allows for the comparison and evaluation of alternative
  technologies. The first release of GATE is now available.
\end{abstract}

\section{Introduction}
\label{sec:intro}
\vspace*{-1ex}

This paper reviews the currently available design strategies for software
infrastructure for NLP and presents an implementation of a system called
GATE -- a General Architecture for Text Engineering. By {\em software
infrastructure} we mean what has been variously referred to 
in the literature as: software architecture; software support tools; language
engineering platforms; development environments. Our gloss on these terms is:
{\em common models for the representation, storage and exchange of data in and
between processing modules in NLP systems, along with graphical
interface tools for
the management of data and processing and the visualisation of data.}
NLP systems produce information about texts%
\footnote{These texts may sometimes be the results of automatic speech
recognition -- see section \ref{sec:intarc}.},
and existing systems that aim to provide software infrastructure for NLP
can be classified as belonging to one of three types according to the way in
which they treat this information:
\begin{description}
\item[additive, or markup-based:] information pro\-duc\-ed is
added to the text in the form of markup, e.g.\ in SGML \cite{Tho96};
\item[referential, or annotation-based:] information is stored separately
with references back to the original text, e.g.\ in the TIPSTER architecture
\cite{Gri96b};
\item[abstraction-based:] the original text is preserved in processing only
as parts of an integrated data structure that represents information
about the text in a uniform theoretically-motivated model, e.g.\
attribute-value structures in the ALEP system \cite{Sim94}.
\end{description}
A fourth category might be added to cater for those systems that provide
communication and control infrastructure without addressing the
text-specific needs of NLP (e.g.\ Verbmobil's ICE architecture \cite{Amt95}).

We begin by reviewing examples of the three
approaches we sketched above (and a system that falls into the fourth
category).
Next we discuss current trends in the field and motivate a set of
requirements that have formed the design brief for GATE, which is then
described.
The initial
distribution of the system includes a MUC-6 (Message Understanding
Conference 6 \cite{Gri96}) style information
extraction (IE) system and an overview of these modules is given.
GATE is now available for research purposes -- see \\
\small
{\tt http://www.dcs.shef.ac.uk/research/groups/ nlp/gate/}
\normalsize
for details of how to obtain the system. It is written in C++ and Tcl/Tk and
currently runs on UNIX (SunOS, Solaris, Irix, Linux and AIX are known to
work); a Windows NT version is in preparation.

\section{Managing Information about Text}
\label{sec:managing}
\vspace*{-1ex}

\subsection{Abstraction Approaches}

The abstraction-based approach to managing information about texts is
primarily motivated by theories of the nature of the
information to be represented. One such position is that declarative,
constraint-based representations using
feature-structure matrices manipulated under unification
are an appropriate vehicle by which ``many technical problems in language
description and computer manipulation of language can be solved''
\cite{Shi92}. Information in these models may be characterised as {\em
abstract} in our present context as there is no requirement to tie data
elements back to the original text -- these models represent abstractions
from the text.

One recent example of an infrastructure project based on abstraction
is ALEP -- the Advanced Language Engineering Platform
\cite{Sim94}.
ALEP aims to provide ``the NLP research and engineering
community in Europe with an open, versatile, and general-purpose development
environment''.
ALEP, while in principle open, is primarily
an advanced system for developing and
manipulating feature structure knowledge-bases under unification. Also
provided are several parsing algorithms, algorithms for transfer, synthesis
and generation \cite{Sch94}. As such, it is a system for developing particular
types of data resource (e.g.\ grammars, lexicons) and for {\em doing} a
particular set of tasks in LE in a particular way. ALEP does not aim for
complete genericity (or it would need also to supply algorithms for Baum-Welch
estimation, fast regular expression matching, etc.). Supplying a generic
system to do every LE task is clearly impossible, and prone to instant
obsolescence in a rapidly changing field.

In our view ALEP, despite claiming to use a theory-neutral formalism (an
HPSG-like formalism), is still too committed to a particular approach to
linguistic analysis and representation. It is clearly of high utility to those
in the LE community to whom these theories and formalisms
are relevant; but it
excludes, or at least does not actively support, all those who are not,
including an increasing number of researchers committed to statistical,
corpus-based approaches. GATE, as will be seen below, is more like a shell, a
backplane into which the whole spectrum of LE modules and databases can be
plugged. Components used within GATE will typically exist already -- our
emphasis is reuse, not reimplementation. Our project is to provide a flexible
and efficient way to combine LE components to make LE systems (whether
experimental or for delivered applications) -- not to provide `the one true
system', or even `the one true development environment'. Indeed, ALEP-based
systems might well provide components operating within GATE. Seen  this way,
the ALEP enterprise is orthogonal to ours --  there is no significant overlap
or conflict.

In our view the level at which we can assume commonality of
information, or of representation of information, between LE modules
is very low, if we are to build an environment which is broad enough
to support the full range of LE tools and accept that we cannot impose
standards on a research community in flux. What \emph{does} seem to be
a highest common denominator is this: modules that process text, or
process the output of other modules that process text, produce further
information about the text or portions of it. For example,
part-of-speech tags, phrase structure trees, logical forms, discourse
models can all be seen in this light. It would seem, therefore, that
we are on safe common ground if we start \emph{only} by committing to
provide a mechanism which manages arbitrary information about text.

There are two methods by which this may be done. First, one may embed
the information in the text at the relevant points -- the {\em additive}
approach.  Second, one
may associate the information with the text by building a separate
database which stores this information and relates it to the text using
pointers into the text -- the {\em referential} approach.
The next two subsections discuss systems that
have adopted these two approaches respectively, then
we compare the two and indicate why we have chosen a hybrid
approached based mainly on the second. Finally we look at a system that
falls outside our three categories.

\subsection{Additive Approaches}
\vspace*{-1ex}

Additive architectures for managing information about text add markup to the
original text at each successive phase of processing. This model has been
adopted by a number of projects including parts of the 
MULTEXT EC project. The MULTEXT
work%
\footnote{Note that other partners in the project adopted a different
architectural solution -- see
{\tt http://www.lpl.univ-aix.fr/projects/multext/}.}
has led to the development of an architecture based on SGML
at the University of Edinburgh called LT-NSL \cite{Tho96}.

The architecture is based on a commitment to TEI-style (the Text
Encoding Initiative \cite{Spe94}) SGML encoding of
information about text. The TEI defines standard tag sets for a range of
purposes including many relevant to LE systems. Tools in a LT-NSL system
communicate via interfaces specified as SGML document type definitions (DTDs
-- essentially tag set descriptions), using character streams on pipes -- an
arrangement modelled after UNIX-style shell programming.
To obviate the need to deal with some difficult types of
SGML (e.g. minimised markup) texts are converted to a normal form before
processing.  A tool selects
what information it requires from its input SGML stream and adds information
as new SGML markup. An advantage here is a degree of data-structure
independence: so long as the necessary information is present in its input, a
tool can ignore changes to other markup that inhabits the same stream --
unknown SGML is simply passed through unchanged (so, for example, a semantic
interpretation module might examine phrase structure markup, but ignore POS
tags). A disadvantage is that although graph-structured data may be expressed
in SGML, doing so is complex (either via concurrent markup, the specification
of multiple legal markup trees in the DTD, or by rather ugly nesting tricks to
cope with overlapping -- so-called ``milestone tags''). Graph-structured
information might be present in the output of a parser, for example,
representing competing analyses of areas of text.

\subsection{Referential Approaches}
\label{subsec:tipster}
\vspace*{-1ex}

The ARPA-sponsored TIPSTER programme in the US, now entering its third phase,
has also produced a data-driven architecture for NLP systems \cite{Gri96b}.
Whereas in LT-NSL all information about a text is encoded in SGML, which is
added by the modules, in TIPSTER a text remains unchanged while information is
stored in a separate database -- the referential approach.
Information is stored in the database in the form of {\em
annotations}. Annotations associate arbitrary information ({\em attributes}),
with portions of documents (identified by sets of start/end byte offsets or
{\em spans}). Attributes may be the result of linguistic analysis, e.g.\ POS
tags or textual unit type. In this way the information built up about a text
by NLP modules is kept separate from the texts themselves. In place of
an SGML DTD, an {\em annotation type declaration} defines the information
present in annotation sets.
Figure \ref{annotations_eg} shows an example
from \cite{Gri96b}.
%
\small
\begin{figure}[!htbp]
\small
\begin{center}
\begin{tabular}{|l|l|r|r|l|} \hline
\multicolumn{5}{|c|}{\em Text} \\
\multicolumn{5}{|c|}{{\tt Sarah savored the soup.}} \\
\multicolumn{5}{|c|}{{\tt 0...|5...|10..|15..|20}} \\ \hline \hline
\multicolumn{5}{|c|}{\em Annotations} \\
Id & Type  & \multicolumn{2}{c|}{Span} & Attributes \\
   &       & Start & End &                          \\ \hline
 1 & token & 0  & 5  & pos=NP \\
 2 & token & 6  & 13 & pos=VBD \\
 3 & token & 14 & 17 & pos=DT \\
 4 & token & 18 & 22 & pos=NN \\
 5 & token & 22 & 23 &  \\ \hline
 6 & name  & 0  & 5  & name\_type=person \\ \hline
 7 & sentence & 0 & 23 & \\ \hline
\end{tabular}
\end{center}
\caption{TIPSTER annotations example \label{annotations_eg}}
\end{figure}
\normalsize

The definition of annotations in TIPSTER forms part of an object-oriented
model that deals with inter-textual information as well as single texts.
Documents are grouped into {\em collections}, each with a database storing
annotations and document attributes such as identifiers, headlines etc.
The model also describes elements of information extraction
(IE) and information retrieval
(IR) systems relating to their use,
with classes representing queries and information needs.

The TIPSTER architecture is designed to be portable to a range of operating
environments, so it does not define implementation technologies. Particular
implementations make their own decisions regarding issues such as parallelism,
user interface, or delivery platform. Various implementations of TIPSTER
systems are available, including one in GATE.

\subsection{Comparison of LT-NSL and TIPSTER}
\vspace*{-1ex}

Both architectures are appropriate for NLP, but there are a
number of significant differences. We discuss five here, then note the
possibility of complimentary inter-operation of the two.

\begin{enumerate} \itemsep=0in
\item
TIPSTER can support documents on read-only media (e.g.\ Internet material, or
CD-ROMs, which may be used for
bulk storage by organisations with large archiving needs) without copying
each document.
\item
From the point of view of efficiency, the original
LT-NSL model of interposing SGML between all modules implies a generation
and parsing overhead in each module. Later versions have replaced this model
with a pre-parsed representation of SGML to reduce this overhead. This
representation will presumably be stored in intermediate files, which
implies an overhead from the I/O involved in continually reading and writing
all the data associated with a document to file. There would seem no reason
why these files should not be replaced by a database implementation,
however, with potential performance benefits from the ability to do I/O on
subsets of information about documents (and from the high level of
optimisation present in modern database technology).
\item
A related issue is storage overhead. TIPSTER is minimal in this respect, as
there is no inherent need to duplicate the source text
and all its markup during the nromalisation process.
\item
At first thought texts may appear to be one-dimensional, consisting of a
sequence of characters. This view breaks down when structures like tables
appear -- these are inherently two-dimensional and their representation and
manipulation is much easier in a referential model like TIPSTER than in an
additive model like SGML because a markup-based representation is based on
the one-dimensional view. In TIPSTER,
the column of a table can be represented as a
single object with multiple references to parts of the text (an {\em
annotation} with multiple {\em spans}). Marking columns in SGML requires a
tag for each row of the column.
Related points are that: TIPSTER
avoids the difficulties referred to earlier of representing graph-structured
information in SGML; LT NSL is inefficient where processing algorithms
require non-sequential access to data \cite{McK97a}.
\item
TIPSTER can easily
support multi-level access control via a database's protection mechanisms --
this is again not straightforward in SGML.
\item
Distributed control is easy to implement in a database-centred system like
TIPSTER -- the DB can act as a blackboard, and implementations can take
advantage of well-understood access control (locking) technology. How to do
distributed control in LT-NSL is not obvious.
We plan to provide this type of control in GATE via collaboration with the
Corelli project at CRL, New Mexico -- see \cite{Zaj97b} for more details.
\end{enumerate}

\subsection{Combining Addition and Reference}

We believe the above comparison demonstrates that there are significant
advantages to the TIPSTER model and it is this model that we have chosen for
GATE.

We also believe that SGML and the TEI must remain central to any
serious text processing strategy. The points above do not contradict this
view, but indicate that SGML should not form the central representation format
of every text processing system. Input from SGML text and TEI conformant
output are becoming increasingly necessary for LE applications as more and
more publishers adopts these standards. This does not mean, however, that
flat-file SGML is an appropriate format for an architecture for LE systems.
This observation is born out by the facts that TIPSTER started with an SGML
architecture but rejected it in favour of the current database model, and that
LT-NSL has gone partly towards this style by passing pre-parsed SGML between
components.

Interestingly, a TIPSTER referential
system could function as a module in an LT-NSL additive
system, or vice-versa. A TIPSTER storage system could write data in SGML for
processing by LT-NSL tools, and convert the SGML results back into native
format. Work is underway to integrate the LT-NSL API with GATE and
provide SGML I/O for TIPSTER (and we acknowledge valuable assistance from
colleagues at Edinburgh in this task).

\subsection{ICE}
\label{sec:intarc}
\vspace*{-1ex}

ICE, the Intarc Communication Environment \cite{Amt95}, is an `environment
for the development of distributed AI systems'. As part of the Verbmobil
real-time speech-to-speech translation project ICE has
addressed two key problems for this type of system, viz.\ distributed
processing and incremental interpretation \cite{Gor96}: distribution to
contribute to processing speed in what is a very compute-intensive
application area; incremental interpretation both for speed reasons and to
facilitate feedback of results from downstream modules to upstream ones
(e.g.\ to inform the selection of word interpretations from phone lattices
using part-of-speech information).

ICE provides a distribution and communication layer based on PVM (Parallel
Virtual Machine). The infrastructure that ICE delivers doesn't fit into our
tripartite classification because the communication channels do not use data
structures specific to NLP needs, and because data storage and text
collection management is left to the individual modules.

ICE might well form a useful backbone for an NLP infrastructure, and could
operate in any of the three paradigms.

\section{NLP Trends and GATE}
\label{sec:motivation}
\vspace*{-1ex}

For a variety of reasons NLP
has recently spawned a related engineering discipline called
\emph{language engineering} (LE), whose orientation is towards the
application of NLP techniques to solving large-scale, real-world
language processing problems in a robust and predictable way. These
problems include information extraction, text summarisation,
document generation, machine translation, second language learning,
amongst others. In many cases, the technologies being developed are
assistive, rather than fully automatic, aiming to enhance or
supplement a human's expertise rather than attempting to replace it.

The reasons for the growth of language engineering include:
\begin{itemize} \itemsep=0in
\item computer hardware advances which have increased processor speeds
  and memory capacity, while reducing prices;
\item increasing availability of large-scale, language-related,
  on-line resources, such as dictionaries, thesauri, and `designer'
  corpora -- corpora
  selected for representativeness and perhaps annotated with
  descriptive information;
\item the demand for applications in a world where electronic text
  has grown exponentially in volume and availability, and where
  electronic communications and mobility have increased the importance
  of multi-lingual communication;
\item maturing NLP technology which is now able, for some tasks, to
  achieve high levels of accuracy repeatedly on real data.
\end{itemize}

Aside from the host of fundamental theoretical problems that remain to be
answered in NLP, language engineering faces a variety of problems of its own.
Two features of the current situation are of prime importance; they constrain
how the field can develop and must be acknowledged and addressed. First, there
is no theory of language which is universally accepted, and no computational
model of even a part of the process of language understanding which stands
uncontested. Second, building intelligent application systems, systems which
model or reproduce enough human language processing capability to be useful,
is a large-scale engineering effort which, given political and
economic realities, must
rely on the efforts of many small groups of researchers, spatially and
temporally distributed, with no collaborative master plan.

The first point means that any attempt to push researchers into a theoretical
or representational straight-jacket is premature, unhealthy and doomed to
failure. The second means that no research team alone is likely to have the
resources to build from scratch an entire state-of-the-art LE application
system. Note the tension here: the first point identifies a
centrifugal tendency, pushing researchers into ever greater theoretical
diversity; the second, a centripetal tendency forcing them together.

Given this state of affairs, what is the best practical support that can be
given to advance the field? Clearly, the pressure to build on the efforts of
others demands that LE tools or component technologies -- parsers, taggers,
morphological analysers, discourse planning modules, etc, -- be readily
available for experimentation and reuse. But the pressure towards theoretical
diversity means that there is no point attempting to gain agreement, in the
short term, on what set of component technologies should be developed or on
the informational content or syntax of representations that these components
should require or produce.

Our response to these considerations has been to design and implement a
software environment called GATE -- a
General Architecture for Text Engineering
\cite{Cun95,Cun96b}
--  which attempts to meet the following objectives:
\begin{enumerate} \itemsep=0in
\item support information interchange between LE modules at the
  highest common level possible without prescribing theoretical
  approach (though it allows modules which share theoretical
  presuppositions to pass data in a mutually accepted common form);
\item support the integration of modules written in any
  source language, available either in source or binary form, and
  be available on any common platform;
\item support the evaluation and refinement of LE component
  modules, and of systems built from them,  via a uniform, easy-to-use
  graphical interface which in addition offers facilities for visualising
  data and managing corpora.
\end{enumerate}
The remainder of this paper describes the design of GATE.
In section \ref{sec:gate_des}
we detail the design of GATE. Section \ref{sec:vie} illustrates
how GATE can be used by describing how we have taken a pre-existing
information extraction system and embedded it in GATE.
Section \ref{sec:conclus} makes some concluding remarks.

\section{GATE Design}
\label{sec:gate_des}
\vspace*{-1ex}

Corresponding to the three key objectives identified at the end of section
\ref{sec:motivation}, GATE comprises three principal elements:
GDM, the GATE Document Manager, based on the TIPSTER document
manager; CREOLE, a Collection of REusable Objects for Language Engineering: a
set of LE modules integrated with the system; and GGI, the GATE Graphical
Interface, a development tool for LE R\&D, providing integrated access to the
services of the other components and adding visualisation and debugging tools.

Working with GATE, the researcher will from the outset reuse existing
components, and the common APIs of GDM and CREOLE mean only one integration
mechanism must be learnt. As CREOLE expands, more and more modules will be
available from external sources (including users of other TIPSTER systems).


\subsection{GDM}
\vspace*{-1ex}

The GDM provides a central repository
or server that stores all information an LE system generates about the texts
it processes. All communication between the components of an LE system goes
through GDM, which insulates these components from direct contact with each
other and provides them with a uniform API for manipulating the data they
produce and consume.

The basic concepts of the data model underlying the GDM have been explained in
the discussion of the Tipster model in section \ref{subsec:tipster} above. The
TIPSTER architecture has been fully specified \cite{Gri96b} and its
specification should be consulted for further details, in particular for
definitions of the API. The GDM is fully conformant with the core 
document management subset of this specification.

\subsection{CREOLE}
\label{subsec:creole}
\vspace*{-1ex}

All the real work of analysing texts in a GATE-based LE system is done by
CREOLE modules or objects (we use the terms {\em module} and {\em object}
rather loosely to mean interfaces to resources which may be predominantly
algorithmic or predominantly data, or a mixture of both). Typically, a CREOLE
object will be a wrapper around a pre-existing LE module or database -- a
tagger or parser, a lexicon or ngram index, for example. Alternatively,
objects
may be developed from scratch for the architecture -- in either case the
object provides a standardised API to the underlying resources which allows
access via GGI and I/O via GDM. The CREOLE APIs may also be used for
programming new objects.

When the user initiates a particular CREOLE object via GGI (or when a
programmer does the same via the GATE API when building an LE application) the
object is run, obtaining the information it needs (document source,
annotations from other objects) via calls to the GDM API. Its results are then
stored in the GDM database and become available for examination via GGI or to
be the input to other CREOLE objects.

GDM imposes constraints on the I/O format of CREOLE objects, namely that all
information must be associated with byte offsets and conform to the
annotations model of the TIPSTER architecture. The principal overhead in
integrating a module with GATE
is making the components use byte offsets, if they do
not already do so.

\subsection{GGI}
\vspace*{-1ex}

The GGI is a graphical tool that encapsulates the GDM and CREOLE resources in
a fashion suitable for interactive building and testing of LE components and
systems.
The GGI has functions for creating, viewing
and editing the collections of documents which are managed by the GDM and that
form the corpora which LE modules and systems in GATE use as input data.
The GGI also has facilities to display the results of module or system
execution -- new or changed
annotations associated with the document. These annotations
can be viewed either in raw form, using a generic annotation viewer, or in an
annotation-specific way, if special annotation viewers are available. For
example, named entity annotations which identify and classify proper names
(e.g.\ organization names, person names, location names) are shown by
colour-coded highlighting of relevant words; phrase structure annotations
are shown by graphical presentation of parse trees.
Note that the viewers are general for particular types of annotation, so,
for example, the same procedure is used for any POS tag set, Named-Entity
markup etc. (see section \ref{sec:plug_and_play} below). Thus
CREOLE developers reuse GATE data visualisation code with negligible
overhead.

%

\subsection{Plug and Play}\label{sec:plug_and_play}
\vspace*{-1ex}

The process of integrating existing modules into GATE (CREOLEising) has been
automated to a large degree and can be driven from the interface.
The developer is required to produce some C or
Tcl code that uses the GDM TIPSTER API to get information from the database
and write back results. When the module pre-dates integration,
this is called a {\em wrapper} as it encapsulates
the module in a standard form that GATE expects. When 
modules are developed
specifically for GATE they can embed TIPSTER calls throughout their code and
dispense with the wrapper intermediary.
The underlying module can
be an external executable written in any language (the current
CREOLE set includes Prolog, Lisp and Perl programs, for example).

There are three ways to provide the CREOLE wrapper functions. Packages written
in C, or in languages which obey C linkage conventions, can be compiled
into GATE directly as a Tcl package. This is
{\em tight coupling} and is maximally efficient but necessitates
recompilation of GATE when modules change.
On platforms which support shared libraries C-based
wrappers can be loaded at run-time -- {\em dynamic coupling}.
This is also efficient (with a small penalty at load time) and allows
developers to change CREOLE objects and run them within GATE without
recompiling the GATE system.
Wrappers written in Tcl can also be loaded at run-time -- {\em loose
coupling}. There is a performance penalty in comparison with using the C
APIs, but for simple cases this is the easiest integration route.
In each case the implementation of CREOLE services is 
completely transparent to GATE. 

CREOLE wrappers encapsulate information about the preconditions for a module
to run (data that must be present in the GDM database) and post-conditions
(data that will result). This information is needed by GGI, and is provided
by the developer in a configuration file, which also details what sort of
viewer to use for the module's results and any parameters that need passing
to the module. These parameters can be changed from the interface at
run-time, e.g.\ to tell a parser to use a different lexicon.
Aside from the information needed for GGI to provide access to a module,
GATE compatibility equals TIPSTER compatibility -- i.e.\ there will be very
little overhead in making any TIPSTER module run in GATE.

Given an integrated module, all other interface functions happen
automatically. For example, the module will appear in a graph of all modules
available, with permissible links to other modules automatically
displayed, having been derived from the module pre- and post-conditions.

At any point the
developer can create a new graph from a subset of available CREOLE
modules to perform a task of specific interest.

\section{VIE: An Application In GATE}
\label{sec:vie}
\vspace*{-1ex}

To illustrate the process of converting pre-existing LE systems into
GATE-compatible CREOLE sets
we use as an example the creation of VIE (Vanilla
Information Extraction system) from LaSIE (Large-Scale Information
Extraction system)
\cite{Gai95b}, Sheffield's
entry in the MUC-6 system evaluations. LaSIE module
interfaces were not standardised when originally produced and its
CREOLEization gives a good indication of the ease of integrating other LE
tools into GATE. The resulting system, VIE, is distributed with GATE.

\subsection{LaSIE}
\vspace*{-1ex}

LaSIE was designed as a research system for investigating approaches
to information extraction
and to be entered into the MUC-6 conference \cite{Gri96}.
As such it was a standalone system that was aimed at specific tasks and, while
based on a modular design, none of its modules were specifically designed with
reuse in mind, nor was there any attempt to standardise data formats passed
between modules. Modules were written in a variety of programming languages,
including C, C++, Flex, Perl and Prolog. In this regard LaSIE was probably
typical of existing LE systems and modules.
The high-level tasks which LaSIE performed include the four MUC-6
tasks (carried out on {\em Wall Street Journal} articles) -- named entity
recognition, coreference resolution and two template filling tasks.
The system was a pipelined architecture which processes a text
sentence-at-a-time and consists of three principal processing stages:
lexical preprocessing, parsing plus semantic interpretation, and
discourse interpretation.

\subsection{The CREOLEisation of LaSIE}
\vspace*{-1ex}

As described in section
\ref{subsec:creole}, CREOLEisation of existing LE modules involves providing
them with a wrapper so that the modules communicate via the GDM, by accessing
TIPSTER-compliant document annotations and updating them with new information.
The major work in converting LaSIE to VIE involved defining useful module
boundaries, unpicking the connections between them, and then writing wrappers
to convert module output into annotations relating to text spans and to
convert GDM input from annotations relating to text spans back into the
module's native input format.

The complete VIE system comprises ten modules, each of which is a CREOLE
object integrated into GATE. The CREOLEisation took approximately two person
months. The resulting system has all the
functionality of the original LaSIE system.
However, the interface makes it much easier to use. And, of course, it is now
possible to swap in modules, such as a different parser, with significantly
less effort than would have been the case before. For more details of this
process see \cite{Cun96g}.

VIE and its components are being deployed for a number of purposes including
IE in French, German and Spanish. Experience so far indicates that GATE is a
productive environment for distributed collaborative reuse-based software
development.

\section{Concluding Remarks}
\label{sec:conclus}
\vspace*{-1ex}

Of course, GATE does not solve all the problems involved in plugging diverse
LE modules together. There are three barriers to such integration:
\begin{itemize} \itemsep=0in
\item
managing {\em storage and exchange} of information about texts;
\item
incompatibility of {\em representation} of information about texts;
\item
incompatibility of {\em type} of information used and produced by different
modules.
\end{itemize}
GATE provides a solution to the first two of these,
based on the work of the TIPSTER architecture group. Because GATE
places no constraints on the linguistic formalisms or information content used
by CREOLE modules, the latter problem must be solved by dedicated translation
functions -- e.g.\ tagset-to-tagset mapping -- and, in some cases, by extra
processing -- e.g.\ adding a semantic processor to complement a bracketing
parser.

The recent completion of this work means a full assessment of the strengths
and weaknesses of GATE is not yet possible. The implementation of VIE in GATE,
however, provides an existence proof that the original conception is workable.
We believe that the environment provided by GATE will now allow us to make
significant strides in assessing alternative LE technologies and in rapidly
assembling LE prototype systems. Thus, to return to the themes of
section \ref{sec:motivation},
GATE will not commit us to a particular linguistic theory or
formalism, but it will enable us, and anyone who wishes to make use of it,
to build, in a pragmatic way, on the diverse efforts of others.

\section{Acknowledgements}

This work was supported by the UK Engineering and Physical Sciences Research
Council, grant number GR/K25267, and the EC DG XIII  Language Engineering
programme, grant number LE1-2238.


\begin{thebibliography}{}

\bibitem[\protect\citename{Amtrup}1995]{Amt95}
Amtrup, J.W.
\newblock 1995.
\newblock {ICE -- INTARC Communication Environment User Guide and Reference
  Manual Version 1.4}.
\newblock Technical report, University of Hamburg.

\bibitem[\protect\citename{Cunningham, Gaizauskas, and Wilks}1995]{Cun95}
Cunningham, H., R.G. Gaizauskas, and Y.~Wilks.
\newblock 1995.
\newblock {A General Architecture for Text Engineering (GATE) -- a new approach
  to Language Engineering R\&D}.
\newblock Technical Report CS -- 95 -- 21, Department of Computer Science,
  University of Sheffield.
\newblock Also available as {\tt http://xxx.lanl.gov/ps/cmp-lg/9601009}.

\bibitem[\protect\citename{Cunningham \bgroup et al.\egroup }1996]{Cun96g}
Cunningham, H., K.~Humphreys, R.~Gaizauskas, and M.~Stower, 1996.
\newblock {\em {CREOLE Developer's Manual}}.
\newblock Department of Computer Science, University of Sheffield.
\newblock Available at {\tt
  http://www.dcs.shef.ac.uk/research/groups/ nlp/gate}.

\bibitem[\protect\citename{Cunningham, Wilks, and Gaizauskas}1996]{Cun96b}
Cunningham, H., Y.~Wilks, and R.~Gaizauskas.
\newblock 1996.
\newblock {GATE -- a General Architecture for Text Engineering}.
\newblock In {\em Proceedings of the 16th Conference on Computational
  Linguistics (COLING-96)}, Copenhagen, August.

\bibitem[\protect\citename{Gaizauskas \bgroup et al.\egroup }1995]{Gai95b}
Gaizauskas, R., T.~Wakao, K~Humphreys, H.~Cunningham, and Y.~Wilks.
\newblock 1995.
\newblock {Description of the LaSIE system as used for MUC-6.}
\newblock In {\em Proceedings of the Sixth Message Understanding Conference
  (MUC-6)}. Morgan Kaufmann.

\bibitem[\protect\citename{Gorz \bgroup et al.\egroup }1996]{Gor96}
Gorz, G., M.~Kessler, J.~Spilker, and H.~Weber.
\newblock 1996.
\newblock {Research on Architectures for Integrated Speech/Language Systems in
  Verbmobil}.
\newblock In {\em Proceedings of COLING-96, Copenhagen.}

\bibitem[\protect\citename{Grishman}1996]{Gri96b}
Grishman, R.
\newblock 1996.
\newblock {TIPSTER Architecture Design Document Version 2.2}.
\newblock Technical report, DARPA.
\newblock Available at {\tt http://www.tipster.org/}.

\bibitem[\protect\citename{Grishman and Sundheim}1996]{Gri96}
Grishman, R. and B.~Sundheim.
\newblock 1996.
\newblock Message understanding conference - 6: A brief history.
\newblock In {\em Proceedings of the 16th International Conference on
  Computational Linguistics}, Copenhagen, June.

\bibitem[\protect\citename{McKelvie, Brew, and Thompson}1997]{McK97a}
McKelvie, D., C.~Brew, and H.~Thompson.
\newblock 1997.
\newblock {Using SGML as a Basis for Data-Intensive NLP}.
\newblock In {\em Proceedings of the fifth Conference on Applied Natural
  Language Processing (ANLP-97)}.

\bibitem[\protect\citename{Sch\"{u}tz}1994]{Sch94}
Sch\"{u}tz, J.
\newblock 1994.
\newblock {Developing Lingware in ALEP}.
\newblock {\em ALEP User Group News, CEC Luxemburg}, 1(1), October.

\bibitem[\protect\citename{Shieber}1992]{Shi92}
Shieber, S.
\newblock 1992.
\newblock {\em Constraint-Based Grammar Formalisms}.
\newblock MIT Press.

\bibitem[\protect\citename{Simkins}1994]{Sim94}
Simkins, N.~K.
\newblock 1994.
\newblock {An Open Architecture for Language Engineering}.
\newblock In {\em First Language Engineering Convention, Paris}.

\bibitem[\protect\citename{Sperberg-McQueen and Burnard}1994]{Spe94}
Sperberg-McQueen, C.M. and L.~Burnard.
\newblock 1994.
\newblock {\em Guidelines for Electronic Text Encoding and Interchange (TEI
  P3)}.
\newblock ACH, ACL, ALLC.

\bibitem[\protect\citename{Thompson and McKelvie}1996]{Tho96}
Thompson, H.S. and D.~McKelvie.
\newblock 1996.
\newblock {A Software Architecture for Simple, Efficient SGML Applications}.
\newblock In {\em {Proceedings of SGML Europe '96, Munich}}.

\bibitem[\protect\citename{Zajac}1997]{Zaj97b}
Zajac, R.
\newblock 1997.
\newblock {An Open Distributed Architecture for Reuse and Integration of
  Heterogenous NLP Components}.
\newblock In {\em Proceedings of the 5th conference on Applied Natural Language
  Processing (ANLP-97)}.

\end{thebibliography}

\end{document}